\documentclass[10pt, a4paper]{article}
\usepackage{lrec}
\usepackage{xcolor}
\usepackage{bbm}
\usepackage{amssymb}
\usepackage{graphicx}
\usepackage{tabularx}
\usepackage{soul}
\usepackage{amsmath}

\usepackage{epstopdf}
\usepackage[utf8]{inputenc}

\usepackage{hyperref}
\usepackage{xstring}
\definecolor{darkgreen}{rgb}{0.0, 0.2, 0.13}
\definecolor{britishracinggreen}{rgb}{60,179,113}

\title{Identification of primary and collateral tracks in stuttered speech}

\name{Rachid Riad$^{1,2*}$\thanks{*Part of the work done at the Vector Institute during RR’s summer internship}, Anne-Catherine Bachoud-Lévi$^{2}$, Frank Rudzicz$^3$, Emmanuel Dupoux$^{1}$}

\address{
  $1$ CoML/ENS/CNRS/EHESS/INRIA/PSL Research University, Paris, France \\
  $2$ NPI/ENS/INSERM/UPEC/PSL Research University, Cr\'eteil, France  \\
  $^3$ University of Toronto/Vector Institute/St Michaels Hospital/Surgical Safety Technologies Inc, Toronto, Canada\\ 
         rachid.riad@ens.fr, bachoud@gmail.com, frank@spoclab.com, emmanuel.dupoux@gmail.com\\}

\abstract{
Disfluent speech has been previously addressed from two main perspectives: the clinical perspective focusing on diagnostic, and the Natural Language Processing (NLP) perspective aiming at modeling these events and detect them for downstream tasks. In addition, previous works often used different metrics depending on whether the input features are text or speech, making it difficult to compare the different contributions. Here, we introduce a new evaluation framework for disfluency detection inspired by the clinical and NLP perspective together with the theory of performance from  \cite{clark1996using} which distinguishes between primary and collateral tracks. We introduce a novel forced-aligned disfluency dataset from a corpus of semi-directed interviews, and present baseline results directly comparing the performance of text-based features (word and span information) and speech-based  (acoustic-prosodic information). Finally, we introduce new audio features inspired by the word-based span features. We show experimentally that using these features outperformed the baselines for speech-based predictions on the present dataset.  \\ \newline  }

\begin{document}

\maketitleabstract
\section{Introduction}
\label{sec:intro}
Around 6\% percent of spoken words in non-pathological speech are categorised as disfluent \cite{tree1995effects} and this increases with the cognitive load of the speaker \cite{bortfeld2001disfluency,lindstrom2008effect}. Speaking in real time is a demanding activity, subject to cognitive constraints and pragmatic settings. Under time pressure, a word may not be retrieved, part of a sentence may be revised, unfilled and filled pauses may be inserted, words or part of words may be repeated. Some of these deviations can be viewed as the symptoms of sentence planning problems \cite{mcroberts1996role} or as the results of some \textit{strategies} \cite{clark1998repeating} unfolded under speaker's control to signal something. 

Stuttering is a severe case of speech pathology, that interrupts at a higher rate the flow of speech than in typical Speech Production. Indeed, \textit{in addition} to the 'classic' disfluencies, stutterers can produce other forms of disfluency that are both quantitatively and qualitatively distinguishable from typical forms \cite{lickley2017disfluency}.

Speech pathologists need to quantify all the disfluency events for clinical screening but also to assess potential treatments \cite{yaruss1997clinical}. A large number of factors and speaking settings influence stuttering behaviours and occurrences of disfluencies: interlocutor's characteristics (ex: age, relationship with the speaker), conversational settings (at home, at the hospital, at work), speaking tasks (ex: reading, dialogues, descriptions of scenes). The clinical assessment still rely heavily on subjective and one on one evaluation \cite{yaruss2006overall}. An automatic, reliable procedure would provide Speech Pathologists an objective comparison between clinical facilities and treatments. Besides, detecting automatically disfluencies and stuttering symptoms from speech, in different settings, could  unlock in-home assessments and more frequent trainings for patients. 

Two main issues arise from the literature around disfluency detection. The first one is the lack of public pathological annotated datasets. The second issue is the absence of a clear evaluation protocol for the automatic detection of disfluencies. This might be due to the fact that different communities have different applications in mind. Speech pathologists are interested in the automatic classification of type and duration of disfluencies \cite{yaruss1997clinical}. Most used proprietary datasets, with patients performing reading tasks and where disfluent and non disfluent parts have been balanced \cite{noth2000automatic,yildirim2009automatic,oue2015automatic}. Researchers in the NLP community are  interested in modelling the disfluencies for several reasons \cite{shriberg2001errrr}: text normalisation for downstream tasks such as Dependency Parsing or Semantic Role Labeling. In addition, NLP researchers are also interested in disfluency detection for affective computing \cite{tian2015emotion} applications. 
They use features derived from transcribed text \cite{honal2003correction} using Shriberg's formalism \cite{shriberg1994preliminaries}, and focus on the detection of disfluencies in non-pathological speech in telephonic conversation using datasets like Switchboard \cite{godfrey1992switchboard}. Yet, the work from \cite{goldwater2010words} demonstrated that words preceding disfluent interruption points also have high error rates for speech recognition systems. Finally, psycholinguists and clinicians are interested in the distribution and type of disfluencies, which could inform on speech and language production systems \cite{jackson2015responses} \cite{fromkin1971non} as well as diagnosis. Obviously, for this kind of application, running interview-like speech with minimal annotations would be preferable. Since several hybrid text/speech systems have been proposed \cite{tran2018parsing,yildirim2009automatic}, we believe that a common evaluation method would be beneficial to bridge the gap between these research communities. 

 First,  we introduce a new framework for the evaluation of disfluency detection which would be relevant both for spontaneous and/or pathological speech using metrics combining insights from both NLP and Speech Technologies (ST) communities (Section 2). Second, we test these metrics on a new dataset obtained by force-aligning an annotated corpus of pathological speech (Section 3). All annotations in Praat format and code for evaluation will be released on the GitHub repository of the first author RR
 \footnote{
\tiny \url{https://github.com/Rachine/}}.
Third, we compare the performance of different baseline systems across textual and speech inputs on this dataset that were usually used in NLP and ST (Section 4).
Four, to bridge the gap in performance between NLP and ST methods, we introduce new audio features that improve on the different frame-based baselines (Section 5).

\begin{figure}[!t]
	\centering
	\includegraphics[width=\linewidth]{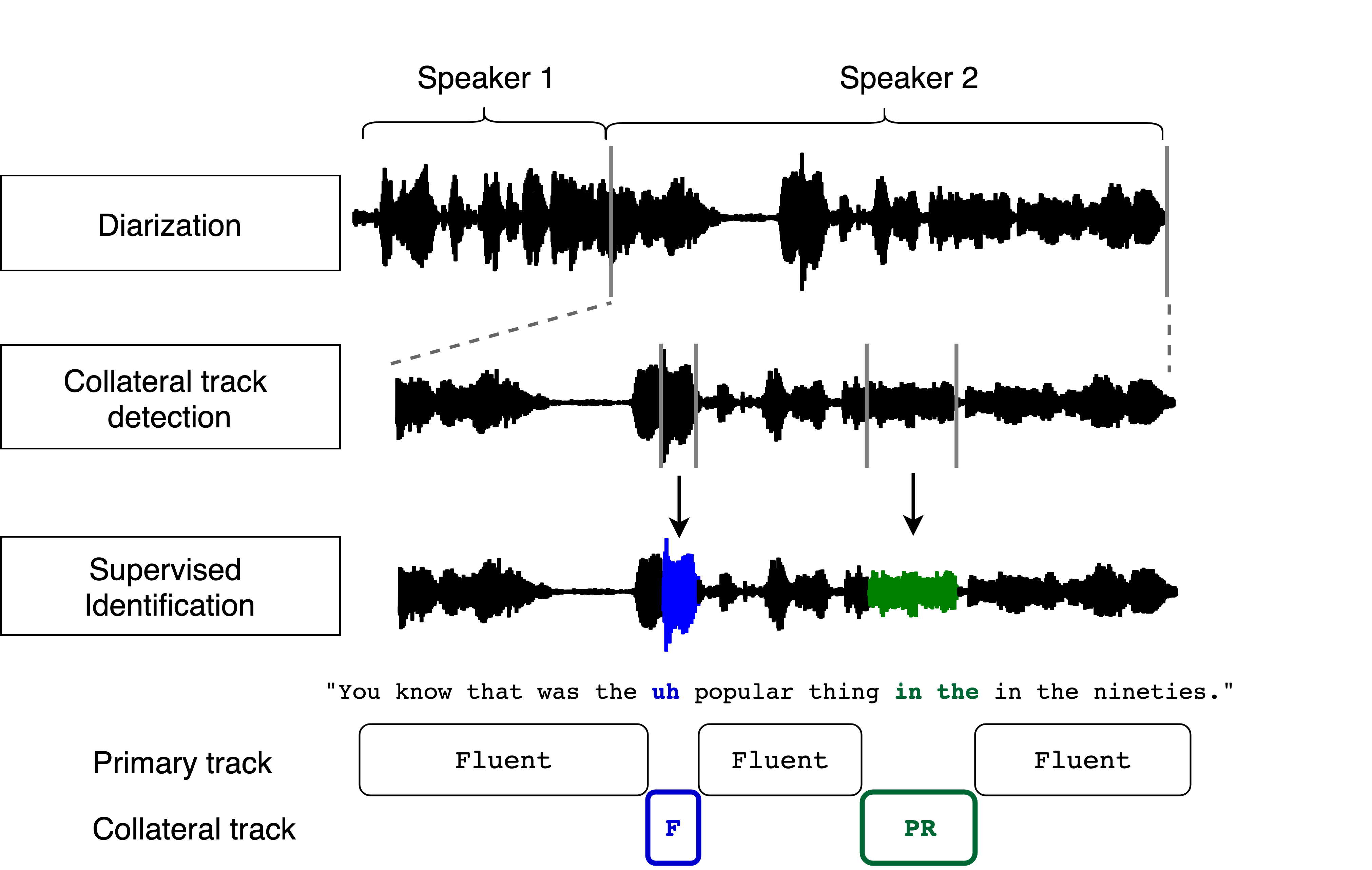}
	\caption{Schematic diagram of identification of the primary and collateral tracks of communication. In black the primary track (fluent), in other colors the words in the collateral track, in \textcolor{blue}{blue} a filler, in \textcolor{darkgreen}{green} a phrase repetition.
	}
	\label{fig:speech_production}
\end{figure}

\section{Metrics: primary and collateral tracks}

We take inspiration from the H. Clark's theory of speech performance \cite[p.~255]{clark1996using}, which states that speakers communicate using two parallel tracks. The \textit{primary track} contains the traditional linguistic content of the discourse while the \textit{collateral track} contains additional signals regulating the communication channel itself. Among these signals we find delays, (un)filled pauses, rephrasing, mistakes, laughs, or vocal noises. The extraction of these two tracks from continuous speech can be decomposed in several engineering tasks (see Figure \ref{fig:speech_production}). First, a diarization \cite{anguera2012speaker} component involves assigning stretches of signal (turns) to each speaker. This component is well studied and existing evaluation metrics can be used (see \cite{bredin2017pyannote} for a suite of diagnostic tools). Second, each turn is analysed in terms which sub-part contains collateral information (collateral detection task). This allows us to quantify how well the collateral track (only  disfluencies in this work) is detected without specifying their category. The parsed segments can be evaluated in terms of a gold lexicon and a gold alignment. For this, we report the Detection Error Rate and the Detection F1-score. Third, each collateral sub-part is categorised into a small number of categories, which we restrict here to disfluency types (identification task). For that purpose we use the Identification F1-score and the Identification Error Rate. The formula to obtain these metrics are summarised in the Table \ref{tab:metrics_detection}. The two error rate metrics are defined with the similar formula to that used in the Diarization Error Rate, which means that they can go over 100\% (the denominator being restricted here to the collateral track). All these metrics were coded using the python toolkit $pyannote.metrics$ \cite{bredin2017pyannote}. 

\begin{table}[t!]

\small
\centering
\begin{tabular}{lc}
\hline
  Metrics & Formula  \\
  \hline
  
	Detection Precision       & $\frac{T_\text{True Positive} }{T_\text{True Positive} + T_\text{false alarm}}$                                                                \\[0.2cm] 
		Detection Recall          & $\frac{T_\text{True Positive} }{T_\text{True Positive} + T_\text{missed detection}}$                                                                \\[0.2cm] 
		Detection F1-score        & $ 2 \frac{ \text{detection precision} \times \text{detection recall} }{\text{detection precision} + \text{detection recall}}$                     \\[0.2cm] 
		Detection Error Rate      & $\frac{T_\text{false alarm} + T_\text{missed detection}}{T_\text{Collateral Track}}$  \\[0.2cm]                                                              
		Identification Precision  & $\frac{1}{5}\sum_{i}\text{Precision}_{i}$                                                              \\[0.2cm] 
		Identification Recall     & $\frac{1}{5}\sum_{i}\text{Recall}_{i}$                                                             \\[0.2cm] 
		Identification F1-score   & $ 2 \frac{ \text{identification precision} \times \text{identification recall} }{\text{identification precision} + \text{identification recall}}$ \\[0.2cm] 
		Identification Error Rate & $\frac{T_\text{false alarm} + T_\text{missed detection} + T_\text{confusion}}{T_\text{Collateral Track}}$                                         
\end{tabular}
\label{tab:metrics_detection}
	\caption{Metrics used for the detection and identification of the collateral track. $T_\text{false alarm}$ is the duration of false alarm (e.g. primary track classified as collateral), $T_\text{miss detection}$ is the duration of missed detection (e.g. collateral track classified as primary), $T_\text{Collateral Track}$ is the total duration of the collateral track in the reference, $T_\text{confusion}$ is the total duration of the confusion between disfluency labels. Detection F1-score is computed as there are only two classes (primary and collateral). $\text{Precision}_{i}$ and $\text{Recall}_{i}$ are computed as the detection formula where the positive class is the i-th disfluency Table \ref{tab:disfluency_table}} 
\end{table}

Another motivation for this framework comes from speech pathology research on Stuttering evaluation. Indeed, from these timing prediction of the primary and collateral track, it can be computed automatically the Speech Efficiency Score (SES) introduced in \cite{amir2018speech}. This study demonstrated that this score, which is based on a time-domain analysis is closely equivalent to stuttering severity ratings done by speech pathologists. By solving the diarization task, and the disfluency detection task mentioned above, it is possible to obtain an estimation of the SES (see below the formula for the equivalence between our framework and their notations). 

  \begin{eqnarray*}
     SES & = \frac{T_\text{Primary Track}}{T_\text{Primary Track} + T_\text{Collateral Track} } * 100 \\
      & = \frac{T_\text{Efficient time}}{T_\text{Total time} - T_\text{Silence} } * 100
  \end{eqnarray*}


\section{Dataset}
We built on FluencyBank, a large-scale open source audiovisual dataset primarily used
by clinical researchers to study fluency \cite{bernstein2018fluency}, from which we selected and forced-aligned a consistent subset focused on stuttering. FluencyBank contains a collection of sub-datasets collected by different research groups to study  typical and disordered fluency in infants and adults. We selected the Adult-who-Stutter(AWS) sub-dataset of \footnote{https://fluency.talkbank.org/access/Voices-AWS.html}{FluencyBank}, which contains video recordings focused on patients. We excluded the recordings where the annotation was lacking, and we obtained 22 speaker video interviews (1429 utterances and 24693 words). The original recordings were done while the participants answered questions of the OASES elicitation protocol \cite{yaruss2006overall}, and transcriptions and disfluencies were done in the original dataset at the sentence level. Table \ref{tab:disfluency_table} provides the five classes of disfluencies that we consider here. We provide some examples for each class of patients answering some questions of the protocol. In this work, we did not consider blocks, syllable repetitions or prolongations. Yet, our formulation with the primary and collateral tracks can easily be extended to these disfluencies.

\begin{figure}[t]
	\centering
	\includegraphics[width=\linewidth]{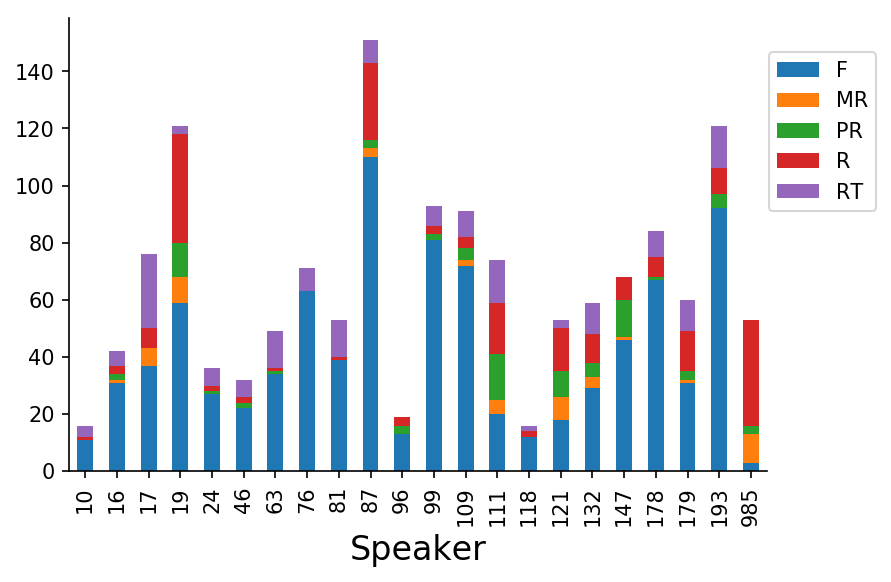}
	\caption{Distribution of disfluency in the FluencyBank-AWS per speaker. See table \ref{tab:disfluency_table} for definitions of disfluencies.}
	\label{fig:dislfuency_stats_aws}
\end{figure}

We obtained the timings of the primary and collateral tracks by force-alignment at the phone level with the kaldi toolkit \cite{povey2011kaldi} with a HMM-GMM model. Figure \ref{fig:dislfuency_stats_aws} shows the distribution of disfluencies per speaker in the dataset. The total number of disfluencies and their types vary greatly across speakers.


\begin{table}[th]
\small
	\caption{Collateral signals taxonomy (usually called disfluency) under consideration here in the FluencyBank dataset:
		\textit{Italic} for the primary track and \textbf{Bold} for the collateral track.}
	\label{tab:disfluency_table}
	\centering
	\begin{tabular}{ll}
	\hline
		\textbf{Disfluency}           & \textbf{Example}                                                            \\
\hline
		Filled pause  (F)             & \textit{I was primarily }\textbf{uh}  \textit{focused} \\ &\textit{on fluency.}           \\
		Single word Repetition    (R) & \textbf{I} \textit{I don't like switch word.  }                             \\
		Multi-Repetition (MR)         & \textit{I'm fortunate to} \textbf{be be be} \\ & \textit{be in graduate school.} \\
		Phrase Repetition   (PR)      & \textbf{they are} \textit{they are so sweet.}                               \\
		Retracing or Revision (RT)    & \textbf{I ended when I was} \textit{it ended} \\ & \textit{ when I was seventeen.}         \\
		\hline
	\end{tabular}
			  
\end{table}

\section{Baselines: text versus speech predictions}
Here, we provide two different kind of baseline systems with the purpose of comparing textual and acoustic approaches on the same metrics: (1) word-based systems, which assume that the input speech has been segmented into words, and aggregate textual and/or acoustic features over the entire span of each word (2), frame-based system which make decision on a frame-by frame basis from raw speech. Obviously, the latter kind of system cannot use textual features. All evaluations are performed with leave-one-speaker-out cross-validation, so that we can assess the generalisation to unseen speaker.

\subsection{Detection from aligned speech: word-based systems}
Word-based systems can incorporate both textual and acoustic features (See Table \ref{tab:disfluency_nlp_features}). 
As for textual features, we used token and span features which are common in the NLP community \cite{shriberg1994preliminaries,charniak2001edit}. As for acoustic/prosodic, we use summary statistics on duration energy and F0. All statistics and pooling are done using the timing alignment of each word $w_i$. The semantic representation and part-of-speech tags are extracted with \cite{spacy2}. The number of syllables and phones are extracted with \cite{mathieu_bernard_2019_2590455}.


\begin{table}[th]
	\caption{List of features in the word-based prediction.}
	\label{tab:disfluency_nlp_features}
	\centering
	\begin{tabular}{|l|c|c|} 
		\hline
		Type  & \multicolumn{2}{c|}{Core features and dimension} \\
		\hline
		\hline
		Token    & semantic representation                   & 384 \\
		\cline{2-3} 
		         & part-of-speech (pos): $p_i$                & 19  \\
		\cline{2-3} 
		         & word position                                           & 1   \\
		\hline
		Span     & $ w_i==w_{i+k},  k \in [-15,+15]^*$                     & 30  \\
		\cline{2-3} 
		         & $p_i==p_{i+k}, k \in [-15,+15]^*$                       & 30  \\
		\cline{2-3} 
		         & $ w_i,w_{i+1}==w_{i+k},w_{i+k+1}  $     &  8 \\
		         		      		         & $ k \in [-4,+4]^* $      &    \\
		\cline{2-3} 
		         & $p_i,p_{i+1}==p_{i+k},p_{i+k+1} $      & 8   \\
		      		         & $ k \in [-4,+4]^* $      &    \\
		\hline
		Acoustic & word duration (s)                                       & 1   \\
		\cline{2-3} 
		Prosodic & number of syllables  & 1   \\
		\cline{2-3} 
		         & number of phones     & 1   \\
		\cline{2-3} 
		         & high, low and total energy                & 3   \\
		        		         & in filterbanks               &    \\
		\cline{2-3} 
		         & F0 mean, std, median, min, max                          & 9   \\
		         & 5\%, 25\%, 75\% \& 95\% percentiles                     &     \\
		\cline{2-3} 
		         & surrounding pause times (s)                             & 2   \\
		\cline{2-3} 
		         & pitch breaks inside              & 3   \\
		      		         & and around word limits              &    \\
		\hline
	\end{tabular}
			  
\end{table}

Such hand-crafted features have been used previously in the literature \cite{zayats2016disfluency,ferguson2015disfluency} and have shown to improve the prediction performance. Indeed, neighbour words and prosodic cues are very informative about the disfluency events \cite{shriberg1994preliminaries}. In \cite{yildirim2009automatic}, they obtained that interrupting points in disfluencies are 98\% associated with a pitch break. We obtained a similar result in the FluencyBank AWS dataset with 95\% of the disfluent boundaries that match with a pitch break in a 100 ms vicinity. All these features are normalized with the $MaxAbsScaler$ from scikit-learn \cite{scikit-learn} to avoid the loss of sparsity (specially in the span features). 

We compared 5 different models (see Table 4). The latest work is focusing on sequence tagging prediction with Recurrent Neural Network architectures \cite{zayats2016disfluency,tran2018parsing}. We compared Forward and Bi-Directional architectures with Long-Short-Term-Memory (LSTM) Networks for disfluency detection and identification. The hidden dimension of the recurrent networks is set at 20. All the experiments are carried using the Adam training procedure with the default parameters \cite{kingma2014adam} and early-stopping on a held-out validation set of 20\% of spoken utterances of the training speakers. In addition, we used a discriminative approach with classical machine learning classifiers. Every word is supposed to be independent: the goal is now to predict each word individually, without its neighbour's representation or prediction. The information listed in Table \ref{tab:disfluency_nlp_features} is already obtained by the aggregation of local and long range information and could be sufficient to make predictions. We compare a standard classic Support Vector Machine with linear kernel (SVM) and a penalty term of $C=0.025$, a L2 regularised logistic regression and a classic deep forward neural network (DNN) with 2 hidden layers with 100 and 50 hidden units.

In the results, we also report the Token F1 score  \ref{tab:results_global} at the word level to compare the predictions. This differs from the F1-score reported for disfluency detection in the NLP community \cite{godfrey1992switchboard}. Usually, NLP models tries to detect the disfluency and identify the subparts of each disfluency (reperandum, interegnum, repair), not the different categories of disfluencies.

\subsection{Detection from raw speech: frame-based systems}

Frame-based systems have no other information than speech. In principle, they are closer to what would be useful for clinical purposes, but obviously, the task is much harder. For comparison purposes we propose 3 baseline frame-based systems. 
Here, we evaluate frame-level prediction for the disfluency detection and identification as in \cite{oue2015automatic}. The patterns of the disfluencies in the speech signal can range from very local phenomenon (filled pauses) to long time-scales (retracing). Here, we investigate the predictions are made every 10 ms, in a bottom-up manner, using only local features. 

Speech represented using a bank of 40 log-energy Mel-scale filters representing 25 ms of speech (Hamming windowed) every 10 ms. The Mel features are mean-variance normalised per file, using the VAD information. Besides, we extract prosodic features with the F0 trajectory and its first derivatives in a 50 ms window (obtaining a 56-dimension vector as F0 is computed every 1.8ms). These spectral and prosodic representation are concatenated to obtain a final 106-dimension vector representation every 10 ms. All the frame-based systems use a window of 7 stacked frames \cite{oue2015automatic}.

Based on these representations directly extracted from the signal, we follow a similar procedure as in the word-based predictions: we compare a standard classic Support Vector Machine with linear kernel (SVM) and a penalty term of $C=0.025$, with a classic deep forward neural network (DNN) with 2 hidden layers with 100 and 50 hidden units. These approaches have been previously used in stuttering detection literature \cite{chee2009overview}. 

\begin{table*}[t]
	\centering
		\caption{Results of the evaluation of detection and identification of primary and collateral track for the different approaches described in Section 4. The best scores for each metric for each condition (word vs frame based) are in \textbf{bold}, best metrics overall are \underline{underlined}. For the evaluation of the Audio Span Features, we report the performance  with a DNN model trained with the Standard sampler.}
	\begin{tabular}{p{0.22\linewidth}|p{0.05\linewidth}|p{0.05\linewidth}p{0.067\linewidth}p{0.067\linewidth}p{0.06\linewidth}|p{0.05\linewidth}p{0.067\linewidth}p{0.067\linewidth}p{0.05\linewidth}} 
		\hline
		Model & NLP & \multicolumn{4}{c}{Detection} & \multicolumn{4}{|c}{Identification}\\
		\hline
		                            & Token F1 & P     & R     & F1    & Error Rate & P     & R     & F1    & Error Rate \\
		\hline
		\hline
		\multicolumn{9}{l}{\textit{Word-based 4.1 (all features)}}  \\
		
		Forward LSTM                & 0.416  & 0.823 & 0.595 & 0.691 & 0.623      & 0.717 & 0.518 & 0.601 & 0.701      \\
		
		Bi-LSTM                     & 0.417  & 0.786 & 0.605 & 0.684 & 0.731      & 0.701 & 0.537 & 0.608 & 0.799      \\
		
		SVM-Linear                  & \textbf{0.569}  & \underline{\textbf{0.966}} & 0.642 & \underline{\textbf{0.771}} & \underline{\textbf{ 0.381 }}     & \underline{\textbf{ 0.905}} & \underline{\textbf{0.599}} & \underline{\textbf{ 0.721}} & \underline{\textbf{0.424}}      \\
		
		Logistic Regression         & 0.544  & 0.846 & \textbf{0.645} & 0.732 & 0.513      & 0.762 & 0.576 & 0.656 & 0.581      \\
		DNN & 0.485  & 0.958 & 0.611 & 0.746 & 0.417      & 0.855 & 0.544 & 0.665 & 0.484      \\
		\hline
		\hline
		 \multicolumn{9}{l}{\textit{Frame-based 4.2 (Baseline Signal features only) }}\\
		Standard sampler + DNN & --  & 0.312 & 0.014 & 0.026 & 1.005     & 0.182 & 0.010 & 0.020 & 1.008      \\
		Undersampler + DNN & --  & 0.073 & \underline{\textbf{1.000}} & 0.136 & 15.286     & 0.038 & \textbf{0.520} & 0.069 &15.766     \\
		Standard sampler + SVM & -- & 0.150 & 0.086 & 0.109 & 1.502      & 0.116 & 0.067 & 0.086 & 1.520      \\
		Undersampler + SVM & --  & 0.077 & 0.838 & \textbf{0.140} & 12.529      & 0.025 & 0.288 & 0.048 & 13.079  \\
		\hline
\multicolumn{9}{l}{\textit{Frame-based 5 (Audio Span Features) }}\\
		Audio Span Features + Standard Sampler + DNN& --  &\textbf{ 0.864} & $\leq 1\mathrm{e}{-3}$ & $\leq 1\mathrm{e}{-3}$ & 1.000     & \textbf{0.818} & $\leq 1\mathrm{e}{-3}$ & $\leq 1\mathrm{e}{-3}$  & 1.003     \\
		Audio Span Features + baselines features + Standard Sampler + DNN & -- & 0.488 & 0.063 & 0.112 & \textbf{0.986}      & 0.450 & 0.059 & \textbf{0.105} & \textbf{0.990}     \\
		\hline
	\end{tabular}

	    \label{tab:results_global}

\end{table*}
As in many machine learning problems \cite{JMLR:imbalanced2017}, disfluency datasets have the attribute to be very imbalanced. The number of frames that are labelled fluent exceeds by a large margin all the others classes (92.7\% of the frames are labelled as fluent). We evaluate a random undersampler technique \cite{JMLR:imbalanced2017} that discards randomly a large number of the majority class (here fluent) before training each model. This undersampling strategy has been used in Speech Technologies, yet, systems had not been evaluated on running speech datasets.

\section{Audio Span Features}

We want to improve the frame-based system using information over a long time span and replace the textual features with equivalent ones directly from the raw speech. We introduce here our Audio Span Features. The goal of these features is to obtain similar information as the span features from the word-based systems (Table \ref{tab:disfluency_nlp_features}). 

Our main assumptions for disfluency events are: (1) Repetition-like disfluency events exhibit a common underlying structure property in the frequency domain, (2) filled pauses exhibit specific acoustic correlates with a steady frequency signature \cite{gabrea2000detection}, (3) these filled pauses have usually adjacent unfilled pauses/silences \cite{daly1994acoustic}. 

That is why we posit that  local neighbour-similarities in the frequency domain can approximate the span features for the word comparisons from Table \ref{tab:disfluency_nlp_features}. Besides, different chunk size can also inform on the different type of disfluencies (close and more local similarities are triggered by fillers versus spaced and long range similarities are triggered by repetitions of words). 

Therefore, for every time-step $t$, for a given window-scale $s$, we compute the similarity $\psi(t,s,i)$ of the frequency representation $x_t$ centered on $t$ with its  $i$-th closest neighbours. We compute this similarity with the $N$ previous neighbours and the $N$ next. The frequency representation $x_t \in \mathbb{R}^{40}$ is still the bank of 40 log-energy Mel-scale filters computed every $\delta = 10 ms$. These neighbours are centered  centered every $t_i^s=t+s\cdot i \cdot \delta$. The scale $s$ is the (odd) number of stacked frames. So we denote by $x_t^s \in \mathbb{R}^{40 \cdot s}$ the concatenation of the $s$ frames around $t$: 

	  \[
	  x_t^s = \begin{pmatrix}
   x_{t-(\frac{s-1}{2})\cdot \delta} \\
   x_{t-(\frac{s-1}{2}+1)\cdot \delta} \\ 
   ...
   \\
   x_{t+(\frac{s-1}{2}-1)\cdot \delta} \\
   x_{t+(\frac{s-1}{2})\cdot \delta}
\end{pmatrix} 
\]

Finally, our Audio Span Features can be computed:

\begin{equation}
\label{similirogram}
\forall i \in [\![-N, N]\!]^*,  \psi(t,s,i) = x_t^s \cdot x_{t_i^s}^s \cdot \frac{1}{n_{s}}
\end{equation} 

\begin{figure}[t]
	\centering
	\includegraphics[width=\linewidth]{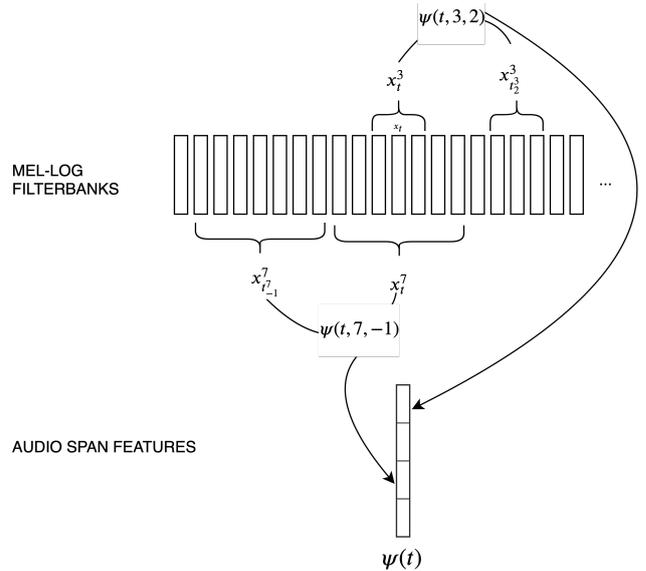}
	\caption{Audio Span Features: It is the concatenation of similarities between the current representation with the $2\cdot N$ closest neighbours filterbank representations at different scales $s$. 
	}
	\label{fig:similirogram}
\end{figure}

We divided the similarity by $n_s=40\cdot s$ to normalise in the scale dimension and not privilege the bigger stacked frames similarities. We computed this similarity for 8 different scales with a logarithmic spacing for the different scales between 30 ms and 1 s ($s \in [101, 61, 37, 23, 13, 9, 5, 3]$). We choose these numbers to capture different orders of magnitude that characterise disfluent segments of speech: phones ($\sim30 ms$), words ($\sim100 ms$), sentences ($\sim1s $).
We finally chose $N=4$ for the number of neighbours ($i \in [-4,-3,-2,-1,1,2,3,4]$). Now, for every time-step $t$ we concatenate the neighbour cross-similarities at different scales and obtain the final vector $\psi(t) \in \mathbb{R}^{4\cdot2\cdot8=64}$. 
See Figure \ref{fig:similirogram} for a schematic representation of the computations.
We evaluate these new Audio Span Features alone and along the acoustic and prosodic representation described in subsection 4.2. We report the evaluation with the Standard Sampler and the DNN model as in 4.2. 


\section{Results and Discussions}

Table \ref{tab:results_global} shows  performances on the detection and identification of primary and collateral tracks. We first review the results from the word-based predictions methods when we take all features as input. Overall, we observed that Sequence-to-sequence models underperform compared to more classical machine learning classifiers. We hypothesise that if the data gets larger the LSTMs architectures might catch on compared to the classifiers. With respect to the F1-score in the detection and identification tasks, the LSTMs architecture are actually not that far from classifiers. Yet, there is an important drop in performance on Error Rates. This highlights the importance to take into account more than one composite score.
Among these classifiers, good old SVM-linear model yields the best performances in almost all metrics (except in the Detection recall for the Logistic Regression). 

Now, we turn to results from the frame-based methods. The results show a sharp drop in performance for all the systems in comparison to the word-based predictions. With the standard sampler, both the DNN and SVM are missing a large number of the disfluency events (Detection Recall at 0.014 and 0.086 respectively). The undersampling technique improves by a large margin the Detection Recall. By contrast, the Precision metrics, the Detection Error Rate and Identification Error Rate are way above 100\%. This shows that with extreme class imbalance, frame-based methods that were previously shown reasonable performance in balanced datasets fail in a spectacular fashion. This highlights the importance of addressing the issue of the detection of disfluencies using running speech rather than artificially balanced datasets or read-speech.

The new Audio Span Features alone demonstrate really poor performance and are missing almost all the disfluency events (Detection Recall and F1 lower than $ 0.001$). However, the Audio Span Features along the acoustic and prosodic representation show the best performance on the frame-based system, especially in the identification task (Identification F1 0.118 and Error Rate below 1). The system misses a number of disfluent events (Low Detection Recall 0.063), but maintain a good precision level in comparison to the other frame-based baselines (Detection Precision 0.488 and Identification Precision 0.450). The Audio Span Features do not capture all the necessary information and can be improved. Especially, our Audio Span Features do not have grammatical information captured by the word-based span features. 
One of our hypothesis is that the Audio Span Features fail also to detect the revision/retracing disfluent events.

\begin{table}[ht]
	\caption{Results of the evaluation of detection and identification of primary and collateral track for the different input features in the word-based predictions with a SVM model with linear kernel. The best error rates for each metric overall are in \textbf{bold}.}
	\centering
	\begin{tabular}{p{0.25\linewidth}|p{0.15\linewidth}p{0.1\linewidth}|p{0.15\linewidth}p{0.1\linewidth}p{0.05\linewidth}p{0.13\linewidth}} 
		\hline
		
		Features & \multicolumn{2}{c}{Detection} & \multicolumn{2}{|c}{Identification}\\
		\hline
		
		                                & F1    & Error Rate     & F1    & Error Rate \\
		\hline
		\hline
		\multicolumn{5}{l}{\textit{Word-based (SVM-Linear)}}\\ 
		Token only                & 0.639 & 0.537     & 0.622 & 0.550      \\
		
		Span only             & 0.313 & 0.826       & 0.263 & 0.856      \\
		
		Acoustic  only   & $\leq 0.001$ & 1.000     & $\leq 0.001$ & 1.000      \\
					Acoustic+Span     & 0.314 &	0.825 &  0.269&0.852  \\
					Acoustic+Token     & 0.646	& 0.529    & 0.628  &    0.543  \\
				Token+Span     & \textbf{0.772} &	0.382   &	\textbf{0.720}& 0.427 \\
				All (line 3 from Table \ref{tab:results_global})    & 0.771 &	\textbf{0.381}   &	0.721& \textbf{0.424} \\

		\hline

	\end{tabular}

	    \label{tab:results_word_based_features}

\end{table}

To better understand the impact of the input features, we ran an ablation study for the word-based predictions, see Table \ref{tab:results_word_based_features}. We compare the different combinations of features as defined in Table \ref{tab:disfluency_nlp_features}. 
First, the acoustic/prosodic features are not informative on their own to predict disfluencies. Span are better but still not reach the full model performance.  
They might be more suitable to detect the repetition-like disfluencies but not necessarily for the Filled pauses. Obviously, the token based representations have a clear advantage especially for Filled pauses. 
The acoustic/prosodic features provide a little gain for the span and token representation, but the combination of span and token is already sufficient on its own and gets very close to the combination of all features. 

This study could orient future work to bridge the gap in performance between our frame-based predictions and word-based predictions. Indeed, the semantic and grammatical part-of-speech features play an important part in the good results of the word-based systems. To obtain such features from the signal, we could build an Automatic-Speech-Recognition pipeline suited for Stutterers and obtain the word2vec representations \cite{mikolov2013distributed}. Or we can obtain such semantic information directly from the signal \cite{chung2018speech2vec}. 

		
		
		
		





\section{Conclusions}

In this work, we investigated a framework to evaluate the disfluencies detection system in stuttered speech. First, we prepared and adapted an open dataset of Adult-Who-Stutter used by clinical researchers, for the task of disfluency detection from running speech. We provided a suite of metrics based on the forced alignment, that enables to compare word-based predictions and frame based-predictions. This allows the direct comparison between different type of approaches. Finally, we compared different baselines systems with textual or acoustic input features, and using word- or frame based pooling of information. The word-based systems show superior performance, illustrating the need (1) to improve frame-based aggregation of information over a long time span and (2) replace textual features with equivalent ones that can be derived automatically from raw speech. Finally, we introduced new Audio Span Features that show the best performances for the frame-based methods. 


\section{Acknowledgements}
This work is funded through a Facebook AI Research grant, and supported by INRIA, as well as grants ANR-10-IDEX-0001-02 (PSL*) and ANR-17-EURE-0017 and grants from FacebookAI Research (Research Grant), Google (Faculty ResearchAward) and Microsoft Research (Azure Credits and Grant).
\newpage

\section{Bibliographical References}
\label{main:ref}
\bibliographystyle{lrec}
\bibliography{lrec2020W-xample}


\end{document}